\documentclass[runningheads]{llncs}

\usepackage{eccv}

\usepackage{eccvabbrv}

\usepackage{graphicx}
\usepackage{booktabs}
\usepackage{enumitem}

\usepackage[accsupp]{axessibility}  

\usepackage{hyperref}

\usepackage{orcidlink}

\begin{document}

\title{Sequential Representation Learning via Static-Dynamic Conditional Disentanglement} 

\titlerunning{Static-Dynamic Disentangled Representation Learning}

\author{Mathieu Cyrille Simon\inst{1} \and
Pascal Frossard\inst{2}\orcidlink{0000-0002-4010-714X} \and
Christophe De Vleeschouwer\inst{1}\orcidlink{0000-0001-5049-2929}}

\authorrunning{M.C.~Simon et al.}

\institute{UCLouvain, ICTEAM, Louvain-la-Neuve, Belgium\\
\email{\{mathieu.simon, christophe.devleeschouwer\}@uclouvain.be} \and
EPFL, LTS4 laboratory, Lausanne, Switzerland\\
\email{pascal.frossard@epfl.ch}}

\maketitle

\begin{abstract}
  This paper explores self-supervised disentangled representation learning within sequential data, focusing on separating time-indep- endent and time-varying factors in videos. We propose a new model that breaks the usual independence assumption between those factors by explicitly accounting for the causal relationship between the static/dynamic variables and that improves the model expressivity through additional Normalizing Flows. A formal definition of the factors is proposed. This formalism leads to the derivation of sufficient conditions for the ground truth factors to be identifiable, and to the introduction of a novel theoretically grounded disentanglement constraint that can be directly and efficiently incorporated into our new framework. The experiments show that the proposed approach outperforms previous complex state-of-the-art techniques in scenarios where the dynamics of a scene are influenced by its content. 
\end{abstract}

\section{Introduction}\label{sec:introduction}

\begin{figure}[tb]
\centering
\begin{minipage}{0.6\textwidth}
  \includegraphics[width=1\linewidth]{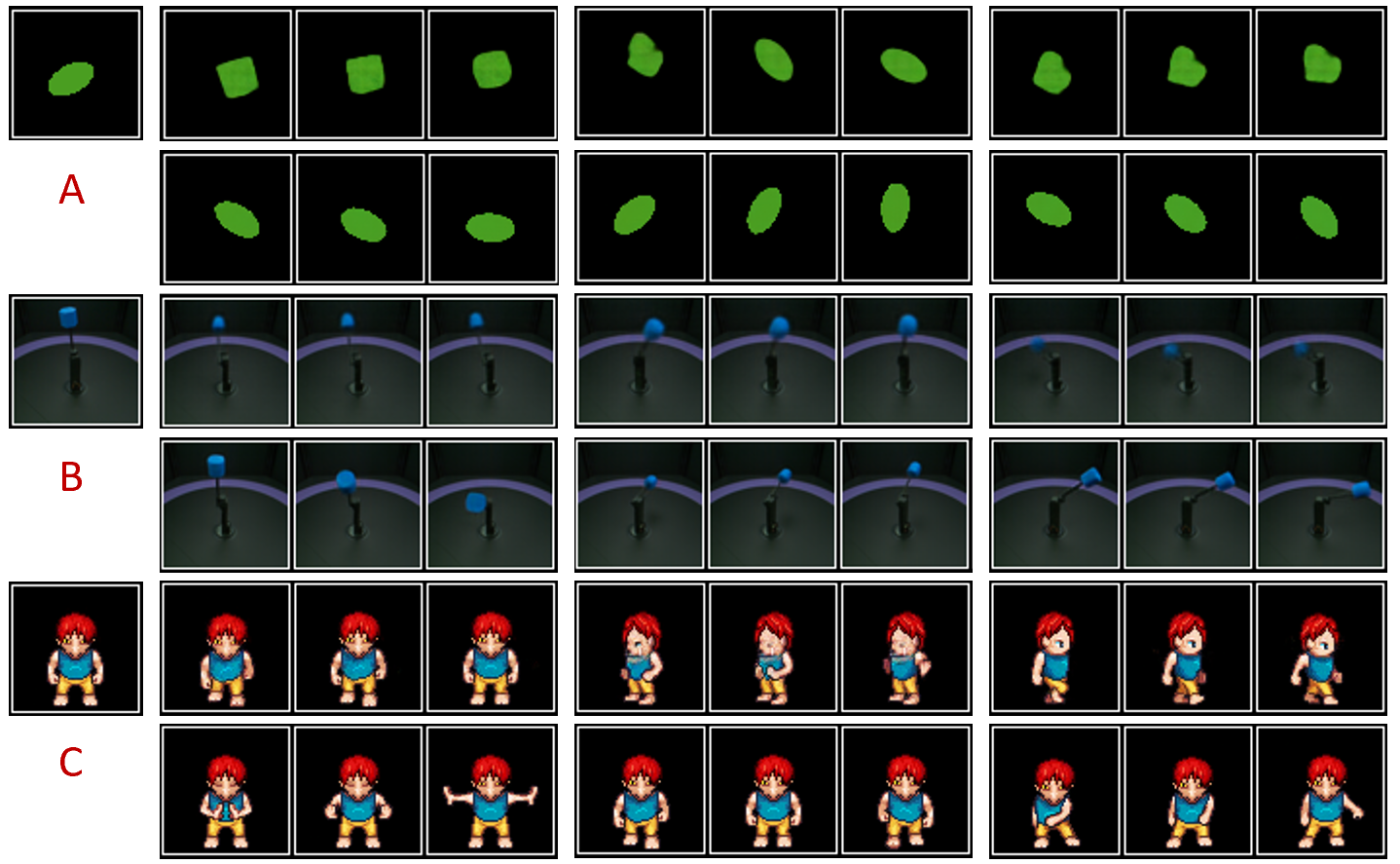}
  \caption{Dynamic generation results for the \textit{c-dSprites} (A), \textit{MPI3D} (B) and \textit{LPCSprites} (C) datasets. The sequences are generated by fixing the static code to the value given by the left image and sampling the dynamic variables from the prior. For each dataset, the first row corresponds to samples from the competing CDSVAE model \cite{bai2021contrastively} and the second row to samples from our proposed model.}
  \label{fig:result1}
\end{minipage}%
\hfill
\begin{minipage}{0.35\textwidth}
  \includegraphics[width=1\linewidth]{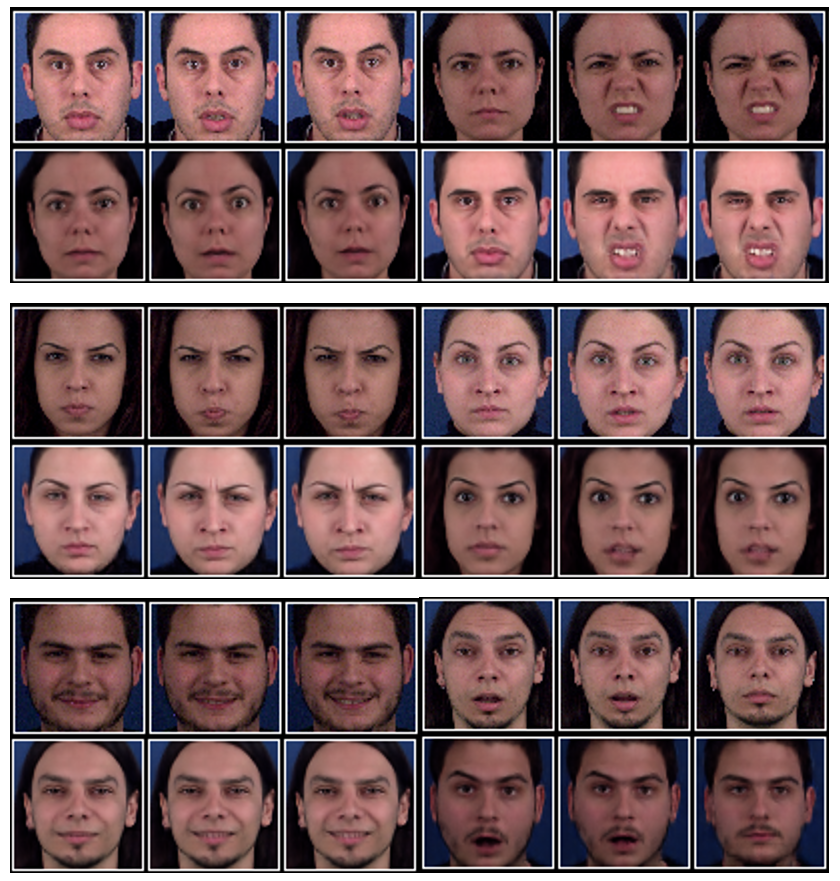}
  \caption{Static/dynamic swap results for the MUG dataset. Odd rows are test input sequences while even rows are sequences generated by swapping the static and dynamic codes of the test sequences using our model.}
  \label{fig:result2}
\end{minipage}
\end{figure}

Disentangled Representation Learning (DRL) focuses on embedding high dimensional complex data into a low dimensional space which factorizes the hidden underlying factors of variations. This disentanglement is expected to facilitate numerous downstream generation or classification tasks and enhance model explainability \cite{locatello2019disentangling,liu2022learning,wang2022disentangled,fragemann2022review,bengio2013representation}. The present work in particular is concerned with unsupervised sequential data DRL \cite{villegas2017decomposing, tian2021good, tulyakov2018mocogan} which, compared to disentanglement on static data \cite{kingma2013auto, wang2022disentangled,zhu2021commutative, locatello2020weakly, bouchacourt2018multi, mita2021identifiable, higgins2016beta}, exhibits specific temporal structure that could be leveraged in the learning process. More precisely, the objective is to learn a representation of video data that factorizes the time independent factors (i.e., what is \textit{static}; the identities, the backgrounds) and time varying factors (i.e., what is \textit{dynamic}; the poses, the motions) in two separate feature vectors. In Fig.\ref{fig:result2}, we show how this disentanglement allows for example to swap the facial expressions (dynamic factors) between faces (static factors). 

As the above description might leave room to some ambiguity into what constitutes a static/dynamic variable, we introduce in this work an explicit and novel formal definition of those factors. Through this formalism, we notably unveil that the factors might not be independent. This paper proposes to investigate the effect of that dependency on the disentanglement models and shows that it translates in difficulties for SotA methods to accurately capture 
\noindent a representation of static and dynamic features. Their extracted codes remain entangled, highligting the importance of developing a robust disentanglement framework that also generalizes to non independent factors. To solve those issues, we propose a new generative model that explicitly incorporates the causal relationship between static and dynamic factors. Diverging from the classical Variational AutoEncoder (VAE) framework, it leverages Conditional Normalizing Flow (cNF) \cite{agrawal2016deep, winkler2019learning, morrow2020variational} to offer improved disentanglement and model expressivity. By exploiting our new formalism, we uncover conditions under which the ground truth factors can be provably inferred and show that our model directly satisfies those conditions using a simple shuffle operation. This leads to a novel evidence lower bound that naturally encourages disentanglement and makes the learned representation provably disentangled.

Among the other works that have addressed this objective under the VAE framework \cite{denton2017unsupervised, hsu2017unsupervised, berman2023multifactor, albarracin2022video}, DSVAE \cite{li2018disentangled} is one of the first models that proposed unsupervised disentanglement of the static/dynamic factors through a factorized sequential latent variable model. However, it was proven to fail in properly disentangling the factors, and to result in representations that heavily depend on the architecture and hyperparameters \cite{luo2022towards}. Haga \etal \cite{haga2023sequential} propose to alleviate this problem using action labels as supervision. However, this requires costly annotations that are rarely available. Subsequent works have therefore extended the DSVAE framework using self-supervised methods aimed at constraining the disentanglement \cite{bai2021contrastively,naiman2023sample,zhu2020s3vae,han2021disentangled,luo2022towards,tonekaboni2022decoupling}. Although these approaches demonstrate improvements over DSVAE, they rely on increasingly complex and cumbersome constraints, besides limiting themselves to cases with independent static/dynamic factors. Comparatively, our approach solves both these problems with our shuffle constraint that does not require any additional loss hence making our method highly simple and non dataset/domain specific. 

\section{Background} \label{sec:background}

Let $\mathcal{G}=\{\mathbf{x}_{1:T_j}^j\}_{j=1}^N$ denote a dataset of $N$ video sequences of length $T_j$ where each $\mathbf{x}^j_t$ corresponds to a video frame at time step $t$ in sequence $j$. The main hypothesis underlying this work is that the frames $\mathbf{x}_t$ can be uniquely represented into two parts, $\mathbf{s}$ and $\mathbf{d}_{t}$ with:
\begin{itemize}[label={$\bullet$}, topsep=0pt, itemsep=0pt]
    \item $\mathbf{s}$ : the \textit{static} factors, common and shared by all frames, representing the \textit{time invariant} information;
    \item $\mathbf{d}_{t}$ : the \textit{dynamic} factors, changing between frames, corresponding to the \textit{time varying} information.
\end{itemize}
Due to temporal coherence, consecutive frames have common factors that are shared between them. Therefore, inspired by Pattern Transformation Manifolds (PTM) representation \cite{vural2012learning}, we consider that each frame resides on a common manifold that corresponds to that shared content. The code $\mathbf{s}$ represents the manifold while $\mathbf{d}_{1:T}$ captures the varying parts specific to each frame. For example, for a video of a person walking, all frames share the same background and person identity thus $\mathbf{s}$ represents those static factors, while $\mathbf{d}_{1:T}$ encodes the person specific pose at each frame i.e., the motion. 

\subsection{Disentangled Sequential VAE}

With the above representation, the work in \cite{li2018disentangled} proposed the following probabilistic model where each frame is generated from its corresponding static and dynamic codes and the dynamic codes are obtained given previous time steps:
\begin{equation}
    p(\mathbf{x}_{1:T},\mathbf{s},\mathbf{d}_{1:T}) = p(\mathbf{s})\textstyle\prod_{t=1}^{T}p(\mathbf{d}_t|\mathbf{d}_{<t})p(\mathbf{x}_t|\mathbf{s},\mathbf{d}_t).
\end{equation}
The objective is to extract the codes $\mathbf{s}$ and $\mathbf{d}_{1:T}$ given the observed data i.e., to learn the factorized posterior distribution $p(\mathbf{s},\mathbf{d}_{1:T}|\mathbf{x}_{1:T})$. Variational inference can be used to learn an approximation of that posterior:
\begin{equation} \label{eq:2}
    q(\mathbf{s},\mathbf{d}_{1:T}|\mathbf{x}_{1:T})= q(\mathbf{s}|\mathbf{x}_{1:T}) \textstyle\prod_{t=1}^T q(\mathbf{d}_{t}|\mathbf{s},\mathbf{d}_{<t},\mathbf{x}_{\leq t})
\end{equation}
with the two terms obtained through separate sequential networks \cite{li2018disentangled}. The model is trained via the VAE algorithm. This encoder network, introduced in \cite{li2018disentangled}, is named 'full q' and is the building block of all following sequential disentanglement works. Unfortunately, despite the disentanglement bias imposed by the factorized latent representation, this model might not be sufficient to achieve a well-disentangled representation. Indeed, as formalized by \cite{locatello2019challenging}, unsupervised disentanglement is fundamentally impossible to achieve without inductive biases since there exists a potentially infinite number of entangled $\mathbf{s}$ and $\mathbf{d}_{1:T}$ having the same marginal $\mathbf{x}_{1:T}$, making the generative model non identifiable. Additionally, VAEs are known to suffer from the so-called 'information preference property' \cite{chen2016variational,zhao2017towards}.  A model with a complex conditional likelihood distribution tends to model a maximum of information in the decoder and ignores the latent variables that become non-informative and trivially match the prior.

DSVAE \cite{li2018disentangled} tries to mitigate that problem using low dimensional dynamic codes $\mathbf{d}_{1:T}$ to promote information in the static variable. However, this requires careful adaptation of the hyperparameters making this method highly impractical without any guarantees on the resulting representation. Following works have therefore extended the DSVAE framework in order to constrain the disentanglement using self-supervised methods that augment the model loss with mutual information (MI) terms. More specifically, the aim is to maximize $I(\mathbf{x}_{1:T},\mathbf{s})$ and $I(\mathbf{x}_{1:T},\mathbf{d}_{1:T})$ while simultaneously minimizing $I(\mathbf{d}_{1:T},\mathbf{s})$. 
Estimating those MI terms is not straightforward and is done either through domain specific methods (e.g., data augmentation, adversarial training or external models \cite{bai2021contrastively,zhu2020s3vae,han2021disentangled}), or through the model itself (e.g., costly encoding-decoding counterfactual losses \cite{tonekaboni2022decoupling} or contrastive MI estimation with VAE-based sampling \cite{naiman2023sample}). Additionally, noting an apparent improvement on disentanglement, these methods modify Eq.(\ref{eq:2}) in order to model both codes independently by ablating the conditional link. 

The above models however have several limitations:
\begin{enumerate}[topsep=0pt, itemsep=0pt]
    \item While the proposed constraints have shown to improve over DSVAE, these methods are either modality-based i.e., data/task-dependent, or they rely on 
    the inherent ability of their model architecture to naturally disentangle. Thus, might end up promoting an improperly disentangled representation.
    \item Even for simple datasets, semantically meaningful factors are generally not independent \cite{yang2020causalvae}. Hence, the conditional link between the two codes should \textit{not} be ablated. We argue that the improvement observed from ablation is due to a lack of expressivity in the conditioning model and MI constraints that are contradicting under potential conditionality. 
    \item The definition of the static and dynamic factors is not formal,
    leaving room to some ambiguity into what constitutes a static/dynamic factor.
\end{enumerate}

\subsection{Other Related Works} 

Differently than the VAE-based methods derived from the DSVAE model \cite{li2018disentangled,naiman2023sample,bai2021contrastively,han2021disentangled,zhu2020s3vae,tonekaboni2022decoupling,albarracin2022video,luo2022towards,haga2023sequential}, Berman \etal \cite{berman2023multifactor} recently developed structured Koopman autoencoders that try to achieve multifactor disentanglement of the sequential data with more than two disentangled components. However, their results still lie behind recent DSVAE-based models \cite{naiman2023sample}. In parallel, the work FHVAE \cite{hsu2017unsupervised,hsu2018scalable} also proposed unsupervised disentanglement through an LSTM-based model but is limited to audio data. 
Then, within the works that try to tackle disentanglement learning for static data \cite{chen2018isolating,kim2018disentangling,higgins2016beta}, we can mention the field of style/content disentanglement \cite{bouchacourt2018multi,von2021self,gabbay2019demystifying,yin2022retriever}, with a definition of the static variables related to ours (Sec.\ref{sec:formalism}).  Interestingly, the effectiveness of those methods can thus be explained through the prism of our formalism in Sec.\ref{sec:method_prop}. This further demonstrates our propositions and justifies our approach for sequential disentanglement, which comparatively does not require additional losses \cite{von2021self,gabbay2019demystifying} or model settings that depend on the dataset \cite{yin2022retriever}.
Finally, several GAN-based models have been successfully applied for disentangling style/content \cite{karras2019style,karras2020analyzing} or appearance/motion for video generation \cite{tulyakov2018mocogan,tian2021good,wang2020g3an}. However, compared to VAE approaches, these methods do not allow to easily encode the input frames and are thus typically less applicable in the context of representation learning.

For image generation, several works have also explored improving VAE models expressivity using NFs to augment either the approximated posterior \cite{kingma2016improved,vahdat2020nvae}, the prior \cite{chen2016variational} or the conditional likelihood \cite{winkler2019learning,agrawal2016deep}. For videos, Marino \etal \cite{marino2020improving} tried to augment the conditional likelihood of sequential latent variable models based on Autoregressive flows \cite{huang2018neural}. However, none of these approaches were investigated for disentanglement. Closer to our work, Ma \etal \cite{ma2020decoupling} decouple the global and local representation of images by embedding a NF in the VAE framework to model the decoder. This shows that cNF naturally tends to disentangle, hence justifying the approach taken in this paper.

Following the work of Locatello \etal \cite{locatello2019challenging}, a handful of methods have recently tried to propose weak-supervision methods based on pairs of images \cite{locatello2020weakly} to ensure the representation from a VAE is identifiable. This was extended to causal variables in \cite{brehmer2022weakly} and to sequences in \cite{lippe2022citris}. However, those methods rely either on knowing the number of changing variables or on labels. 

\textbf{Contributions }: Our work seeks to solve the limitations of previous works by first proposing to \textcolor{red}{formalize the definition of the factors} and then building on this formalism to \textcolor{blue}{introduce a new model that accounts for the potential dependency between the variables}. Our model extends the classical VAE approach through the use of additional Normalizing Flows. In stark contrast to previous works, we show that, by leveraging the data structure through a simple shuffle operation, we can obtain a \textcolor{orange}{provably disentangled representation without any additional loss or supervision} (no data augmentation \cite{bai2021contrastively}, contrastive loss \cite{naiman2023sample}, external models \cite{zhu2020s3vae} or adversarial training \cite{han2021disentangled}), making our method undemanding and modality free, while having theoretical guarantees. This is detailed in the following sections.

\section{Problem Formulation}\label{sec:formalism}
Before trying to disentangle the static and dynamic factors, we propose a new definition of those concepts. This is done through formalizing the video generative process as a latent variable model that comprises two disjoint codes representing the time varying and time invariant factors. We further provide the assumptions underlying this partition.

\textbf{Disentangled video generative model }: We propose a model where the generative process of each frame consists in first drawing a latent code $\mathbf{z}_t$ from its associated probability density $p(\mathbf{z}_t)$ and then obtaining the observations $\mathbf{x}_t$ by sampling from $p(\mathbf{x}_t|\mathbf{z}_t)$. More formally, the frame generative process is:
\begin{align}
    & \mathbf{z}_t\sim p(\mathbf{z}_t), & \mathbf{x}_t = \mathbf{g}(\mathbf{z}_t)
\end{align}
where $p(\mathbf{z}_t)$ is a smooth, continuous density on $\mathcal{Z}$ with full support and $\mathbf{g}:\mathcal{Z}\rightarrow\mathcal{X}$ is a smooth and invertible function with smooth inverse (diffeomorphism) mapping the frame representation space $\mathcal{Z}\subseteq\mathbb{R}^k$ to the observation space $\mathcal{X}\subseteq\mathbb{R}^n$, with usually $n\gg k$.

Based on the static/dynamic disentanglement assumption presented in Sec.\ref{sec:background}, the video sequence frame codes $\mathbf{z}_{1:T}$ can be partitioned into a single invariant static code $\mathbf{s}$ and temporally varying dynamic codes $\mathbf{d}_{1:T}$, i.e., $\mathcal{Z}=\mathcal{S}\times\mathcal{D}$ with $\mathcal{S}\subseteq\mathbb{R}^{n_s}$ and $\mathcal{D}\subseteq\mathbb{R}^{n_d}$, $k=n_s+n_d$:
\begin{align}
    \label{eq:truegene}
     &\mathbf{z}_{1:T}=(\mathbf{s},\mathbf{d}_{1:T}) \sim p(\mathbf{s}) p(\mathbf{d}_1|\mathbf{s})\textstyle\prod_{t=2}^T p(\mathbf{d}_t|\mathbf{s},\mathbf{d}_{<t}), \\ &\mathbf{x}_t = \mathbf{g}(\mathbf{s},\mathbf{d}_t),
\end{align}
with $\mathbf{s}$ corresponding to the first $n_s$ dimensions of $\mathbf{z}_t$, $\mathbf{s}=\mathbf{z}_{t,1:n_s}$. Each frame is generated from its corresponding static and dynamic factors with the dynamic factors obtained given previous time steps \textbf{and} the static factors $\mathbf{s}$. From Eq.\ref{eq:truegene}, it is directly given that $\forall \mathbf{x}_{1:T} \text{ and } t,t'\in\{1,\cdots,T\}$, $\mathbf{g}^{-1}(\mathbf{x}_t)_{1:n_s} = \mathbf{g}^{-1}(\mathbf{x}_{t'})_{1:n_s} = \mathbf{s}$. That is the static factors are invariant for each sequence. As a result, it can be noted that this expression actually matches the formalism presented in the context of style/content disentanglement \cite{von2021self} for static images but it is generalized here to multiple temporally varying views. It ensues that the dynamic factors can be defined based on the same assumptions than the ones adopted for the 'style' factors in \cite{von2021self}: 
\begin{enumerate}[topsep=0pt, itemsep=0pt]
    \item $p(\mathbf{d}_t|\mathbf{s},\mathbf{d}_{<t}) = \delta(\mathbf{d}_{t,A^c_t}-\mathbf{d}_{t-1,A^c_t})p(\mathbf{d}_{t,A_t}|\mathbf{s},\mathbf{d}_{<t})$ with $A_t\subseteq\{1,... n_d\}$ denoting the subset of varying dynamic components at time step $t\ge2$  with associated $p(A_t)$, and $A^c_t$ the complement of $A_t$.
    \item  $\forall i\in\{1,...n_d\}$, $\exists t\in\{2,...T\}$, $A_t\subseteq\{1,... n_d\}$ s.t. $i\in A_t$; $p_{A_t}(A_t)>0$ and for any $(\mathbf{s},\mathbf{d}_{<t})$, $p(\mathbf{d}_{t,A_t}|\mathbf{s},\mathbf{d}_{<t})$ is smooth and fully supported in some open non empty subset containing $\mathbf{d}_{t-1,A_t}$.
\end{enumerate}
The second assumption means that every dynamic variable varies with time. However, as stated by the first assumption, it does not force all dynamic components to change at each time step or for each sequence, echoing the fact that only a subset of the dynamic factors might be fluctuating in a given sequence. 

This video generative model does not make any assumption on the relationship between the factors and allows for arbitrary marginals $p(\mathbf{s},\mathbf{d}_{1:T})$ with potentially complex non trivial statistical relationships between (and within) $\mathbf{s}$ and $\mathbf{d}_{1:T}$. That is $I(\mathbf{s},\mathbf{d}_{1:T})\geq 0$. Overall, this illustrates that, while the static and dynamic factors constitute two separate semantic concepts, they might not be independent. As a result, similarly to \cite{von2021self, locatello2020weakly} for pairs of images, we propose to further extend our description using a Structural Causal Model (SCM) \cite{yang2021causalvae} to express the relation between factors. 

\textbf{Causal model }: From the assumption that $\mathbf{s}$ is invariant throughout the sequences while the $\mathbf{d}_{1:T}$ varies, it results that the static factors may causally influence the dynamic ones but not the opposite. Indeed, if $\mathbf{s}$ would depend on some dynamic variable, following the definition of $\mathbf{d}$ above, this dynamic variable varies and therefore $\mathbf{s}$ would vary as well, contradicting the static invariance. Assuming the dynamic variables undergo temporally coherent perturbations, videos can thus be interpreted as temporally intervened sequences where successive frames result from the causal generative model under soft intervention on the dynamic factors $\mathbf{d}$. This can be formally written as : 
\begin{equation}
    \mathbf{s}=\mathbf{h}_s(\mathbf{\epsilon}_s), \qquad 
     \mathbf{d}_1=\mathbf{h}_d(\mathbf{s},\mathbf{\epsilon}_{d,1}), \qquad\mathbf{d}_t=\mathbf{h}_d'(\mathbf{s},\mathbf{\epsilon}_{d,<t},\mathbf{\epsilon}_{d,t}) \label{eq:causal}
\end{equation}
with $\mathbf{\epsilon}_s$ and $\mathbf{\epsilon}_{d,1:T}$ independent exogenous variables and $\mathbf{h}_s$,$\mathbf{h}_d$,$\mathbf{h}'_d$ functions describing the causal mechanism. Eq.\ref{eq:causal} shows that $\mathbf{s}$ might causally influence the specific values of the dynamic factors but also their possible variations.

By formalizing the video generative process based on static/dynamic factors disentanglement, the link with data augmentation generative process \cite{von2021self} becomes apparent. In the particular case where $T$=2, the two problems are equivalent with the exception that, compared to style/content, the augmentation is not human-designed but is a result of temporal variations in the sequence. This analogy provides an interpretation of videos as augmented views of a scene through temporal augmentation between each frame. Moreover, in \cite{von2021self} an approach is proposed to provably identify the content for style/content disentanglement. The similarities between the two problems thus also hint at the possibility of proposing a similar approach that would provably isolate the static \textbf{and} dynamic factors for sequential disentanglement. We formulate now this idea of provably identifying the different factors. 

\textbf{Definition 1.} \textit{The ground truth factors $\mathbf{s}$ and $\mathbf{\epsilon}_{d,1:T}$ are said to be identified by learned vectors $\mathbf{f}$ and $\mathbf{\lambda}_{1:T}$ if there exist invertible transformations $\mathbf{m}$ and $\mathbf{a}$ s.t. $\mathbf{f}=\mathbf{m}(\mathbf{s})$, $\mathbf{\lambda}_{1:T}=\mathbf{a}(\mathbf{\epsilon}_{d,1:T})$ i.e., $\mathbf{f}$ and $\mathbf{\lambda}_{1:T}$ are reparametrizations of $\mathbf{s}$, $\mathbf{\epsilon}_{d,1:T}$ respectively.} This means that the learned codes $\mathbf{f}$ should capture exactly the static factors $\mathbf{s}$, while $\mathbf{\lambda}_{1:T}$ should encode the dynamic conditionally to $\mathbf{s}$. This is termed as 'conditional disentanglement'. The use of separate notations serves to differentiate the learned codes $\mathbf{f}$ and $\mathbf{\lambda}_{1:T}$ (extracted by the designed network and constraints) from the ground truth factors $\mathbf{s}$ and $\mathbf{\epsilon}_{d,1:T}$ (imposed by the video generative process) which might not be equal in practice. The potential mismatch between the objective and the generative process is avoided by representing the exogeneous dynamic variables $\mathbf{\epsilon}_{d,1:T}$ instead of $\mathbf{d}_{1:T}$. If the static/dynamic factors are independent, $\mathbf{\epsilon}_{d,1:T}$ and $\mathbf{d}_{1:T}$ are equivalent and our formulation becomes therefore a generalization of the objective of Sec.\ref{sec:background} that also accounts for causal dependencies between factors.
\begin{figure*}[tb]
    \centering
  \includegraphics[width=1\linewidth]{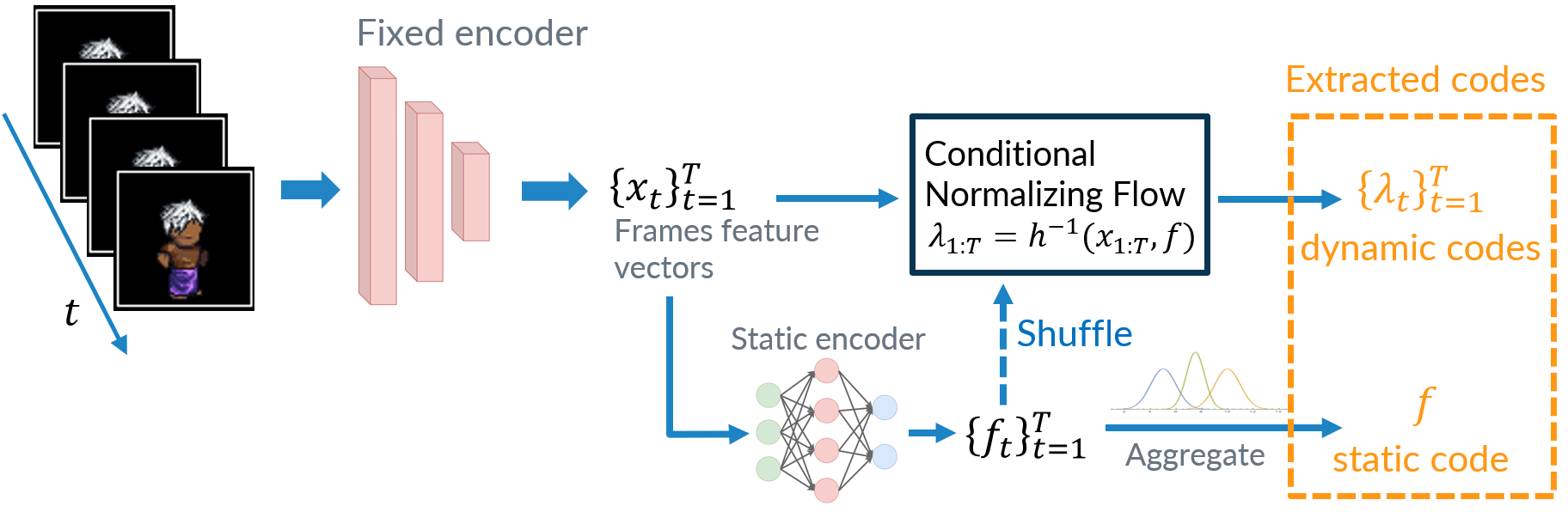}
  \caption{Schematic view of the proposed model. A pretrained Convolutional encoder embeds each frame separately into the frame latent space $\mathbf{x}_{1:T}$. These vectors are used as input to the static encoder which estimates the static codes from each individual frame. The estimations are then aggregated giving the static code $\mathbf{f}$. The static vectors serve to condition a Conditional Normalizing Flow that models the likelihood of the frame feature vectors $\mathbf{x}_{1:T}$. The transformed vectors $\lambda_{1:T}$ correspond to the dynamic codes. The model is trained using the loss $\mathcal{L}$ in Eq.\ref{eq:final}.}
  \label{fig:model}
\end{figure*}

\section{Method} \label{sec:method}

We seek to learn from the video sequences $\mathbf{x}_{1:T}$, two latent codes, a static code $\mathbf{f}$ and dynamic codes $\mathbf{\lambda}_{1:T}$ such that these codes identify the ground-truth factors as presented in Def.1. We start by deriving from our formalism sufficient conditions to provably extract the factors. We then propose a practical model and show how, strictly by leveraging the data structure, this method does achieve the disentanglement conditions. The complete model is summarized in Fig.\ref{fig:model}.

\subsection{Sufficient Conditions for Disentanglement} \label{sec:method_prop}

\textbf{Static disentanglement }: The objective is to learn an encoder posterior distribution $q(\mathbf{f}|\mathbf{x}_{1:T})$ mapping the sequences to their static code such that the static codes $\mathbf{f}$ sampled from $q(\mathbf{f}|\mathbf{x}_{1:T})$ solely capture the static factors $\mathbf{s}$. 

\textcolor{purple}{\textbf{Proposition 1.}} \textit{Consider the generative process and conditions presented in Sec.\ref{sec:formalism}. Further assume that $\text{dim}(\mathbf{f})=n_f\ge n_s$. Let $\mathbf{x}$ and $\mathbf{x}^*$ denote two frames (or disjoint sets of frames) belonging to the same sequence $\mathbf{x}_{1:T}$ and $\mathbb{D}$ be a divergence between two distributions. Given unlimited data from $p(\mathbf{x}_{1:T})$, any candidate posterior distribution $q(\mathbf{f}|\mathbf{x}_{1:T})$ that minimizes 
\begin{equation}
    L_{prop1} = \mathbb{E}_{p(\mathbf{x}_{1:T})}\mathbb{D}(q(\mathbf{f}|\mathbf{x}),q(\mathbf{f}|\mathbf{x}^*))-I(\mathbf{f},\mathbf{s})+I(\mathbf{s},\mathbf{s}),
\end{equation} 
is disentangled in the sense that $q(\mathbf{f})=\int q(\mathbf{f}|\mathbf{x}_{1:T}) p(\mathbf{x}_{1:T}) d\mathbf{x}_{1:T}$, the aggregated posterior, is a reparametrization of $p(\mathbf{s})$.}

The proof is provided in App.A.2. Intuitively, by imposing $\mathbf{f}$ to be both invariant to the dynamic factors and informative of the static ones, the code $\mathbf{f}$ will capture all and only $\mathbf{s}$. Prop.1 extends Thm.4.4 of \cite{von2021self} to distributions and cope with an unknown number of static variables $n_s$. Prop.1 shows that simply maximizing $I(\mathbf{x}_{1:T},\mathbf{f})$ as proposed in Sec.\ref{sec:background} is not sufficient to extract $\mathbf{s}$. Indeed, as shown in \cite{garnelo2018neural,ye2023unified}, a global aggregated code $\mathbf{f}$ could encode the dynamic information and should therefore additionally be made invariant.

\textbf{Dynamic disentanglement }: As explained in Sec.\ref{sec:formalism}, because the static and dynamic factors might not be independent, the goal for the dynamic codes $\mathbf{\lambda}_{1:T}$ is to capture $\mathbf{\epsilon}_{d,1:T}$. Unfortunately, following an approach similar to Prop.1 is not possible in this case. Indeed, this approach would require to have access to two samples $\mathbf{x}$ and $\mathbf{x}^*$ that share the same motion but have different static factors. However, by the definition of the dynamic factors (Sec.\ref{sec:formalism}) there are no easily obtainable such samples since $\mathbf{d}_t$ is specific to each frame and varies at unknown time steps. A solution proposed in \cite{bai2021contrastively} consists in generating positive samples using static data augmentation. However, designing those augmentations might be non trivial and dataset specific, with the risk of artificially and indirectly defining what is static/dynamic through the chosen augmentations. Moreover, only the static part should be affected, limiting the amount of available augmentations, possibly favoring static information in the dynamic codes. Instead, an alternative approach is proposed here that does not require positive samples.

\textcolor{purple}{\textbf{Proposition 2.}} \textit{Consider the same framework as in Prop.1 and the static code $\mathbf{f}$ to be disentangled as described above. Further assume that $\text{dim}(\mathbf{\lambda})=n_\lambda\ge n_d$. Any smooth and invertible candidate function $\mathbf{h}$ s.t. $\mathbf{x}_{1:T}=\mathbf{h}(\mathbf{\lambda}_{1:T},\mathbf{f})$ which minimizes 
\begin{equation}
    L_{prop2} = I(\mathbf{f},\mathbf{\lambda}_{1:T}),
\end{equation}
is disentangled in the sense that $q(\mathbf{\lambda}_{1:T})$ is a reparametrization of $p(\mathbf{\epsilon}_{d,1:T})$.
}

The proof is provided in App.A.3. Intuitively, assuming that $\mathbf{f}$ represents exactly $\mathbf{s}$ thanks to Prop.1 and that both $\mathbf{f}$ and $\mathbf{\lambda}_{1:T}$ are not redundant, because $\mathbf{h}$ is invertible, the dynamic codes $\mathbf{\lambda}_{1:T}$ will have to be informative and to capture all and only $\mathbf{\epsilon}_{d,1:T}$.

Overall, by using the formalism in Sec.\ref{sec:formalism}, we are able to discover sufficient conditions such that the learned codes are reparametrizations of the ground truth factors. We will now show how we learn $q(\mathbf{f}|\mathbf{x}_{1:T})$ and $\mathbf{h}$ in practice so that the conditions are enforced from the observed data only, without requiring explicit formulations of $L_{prop1}$ and $L_{prop2}$.

\subsection{Practical Implementation}

\noindent Compared to previous methods that solely rely on VAEs, we propose instead to improve sequence modeling based on additional Normalizing Flows (NF) \cite{rezende2015variational}. A NF is a transformation of a random variable through a sequence of differentiable invertible mappings. As a result, it is a natural candidate to learn the invertible function $\mathbf{h}$. More specifically, assuming the common static code $\mathbf{f}$ has been extracted from the frames, we model the frames likelihood $p(\mathbf{x}_{1:T}|\mathbf{f})$ using a cNF \cite{winkler2019learning} as
\begin{equation}
    p(\mathbf{x}_{1:T}|\mathbf{f}) = p(\mathbf{\lambda}_{1:T}|\mathbf{f}) \left| \text{det}\frac{\delta \mathbf{h}^{-1}(\mathbf{x}_{1:T},\mathbf{f})}{\delta \mathbf{x}_{1:T}} \right| \label{eq:cNFgzeg}
\end{equation}
with $\mathbf{\lambda}_{1:T} = \mathbf{h}^{-1}(\mathbf{x}_{1:T}, \mathbf{f})$ the dynamic codes. This term can be evaluated through exact likelihood evaluation. Because a NF is a bijective transformation and $\mathbf{f}$ is common to all frames, the transformed latents $\mathbf{\lambda}_{1:T}$ will retain and encode the dynamics of the sequence.  Since the NF is conditioned on the static code $\mathbf{f}$, the dynamics are encoded conditionally to the static part. The NF is based on affine coupling \cite{dinh2014nice} and LSTM layers that model the temporal behavior of the sequence, similarly to \cite{marino2020improving}, and provide a bias towards encoding the dynamics. Because, for a disentangled model, the transformation $\mathbf{h}$ needs to fully model the conditional dependence and causal mechanism,
the final distribution $p(\mathbf{\lambda}_{1:T}|\mathbf{f})$ is replaced with $p(\mathbf{\lambda}_{1:T})$. Details on the architecture are provided in App.B.3.

It remains to show how to extract the static code $\mathbf{f}$ a priori before solving Eq.(\ref{eq:cNFgzeg}). Variational inference is used in order to learn an approximate posterior $q(\mathbf{f}|\mathbf{x}_{1:T})$. Following the standard VAE framework, the marginal data log-likelihood is lower estimated by the Evidence Lower BOund (ELBO):
\begin{align}
     \log p(\mathbf{x}_{1:T}) & \geq \mathbb{E}_{q(\mathbf{f}|\mathbf{x}_{1:T})} \log p(\mathbf{x}_{1:T}|\mathbf{f}) - KL(q(\mathbf{f}|\mathbf{x}_{1:T})||p(\mathbf{f})) \\
     & \geq \mathbb{E}_{q(\mathbf{f}|\mathbf{x}_{1:T})} \log p(\mathbf{\lambda}_{1:T}) + \log \left| \text{det}\frac{\delta \mathbf{h}^{-1}}{\delta \mathbf{x}_{1:T}} \right| - KL(q(\mathbf{f}|\mathbf{x}_{1:T})||p(\mathbf{f})) 
     \label{eq:ELBO}
\end{align}
with the two first terms corresponding to the cNF likelihood. The detailed derivation is shown in App.A.1. Overall, the model can be described as a VAE that extracts a single latent code $\mathbf{f}$ from the observations and has a cNF decoder to model the likelihood of the frames. By maximizing the ELBO, we learn both an approximation of the encoder posterior $q(\mathbf{f}|\mathbf{x}_{1:T})$ and the cNF likelihood i.e., the transformation $\mathbf{h}$. This model does not make any assumption on the independence between the codes and encompasses the potential conditional dependencies through the cNF. In comparison, the 'full q' model (Sec.\ref{sec:background}) is equivalent to model $p(\mathbf{x}_{1:T}|\mathbf{f})$ using a VAE with a latent vector for each time step and a simple concatenation of the codes as inputs to the generative network. The use of a Normalizing flow acts as a more flexible and expressive 'mix-up' of the two extracted codes.

The prior distribution $p(\mathbf{f})$ is further modeled using a NF. This is in contrast with most works that propose to use NF to generate more flexible posteriors instead of priors \cite{kingma2016improved,vahdat2020nvae}. As demonstrated in \cite{chen2016variational}, if these two approaches are equivalent during training, at test time, for sampling, a NF prior allows to have a longer decoder path and produces better samples. To model the inference posterior distribution, the static code encoder $q(\mathbf{f}|\mathbf{x}_{1:T})$ needs to output a single latent code from multiple frames. This is achieved by means of an aggregated posterior distribution \cite{bouchacourt2018multi}. 
The static code distribution $q(\mathbf{f}|\mathbf{x}_{1:T})$ is constructed as the product of individual distributions estimated from each frame by a static encoder network: $q(\mathbf{f}|\mathbf{x}_{1:T})= \prod_{t=1}^{T} q(\mathbf{f}_t|\mathbf{x}_t)$. Compared to previous works where $\mathbf{f}$ is inferred through recurrent layers, the aggregated posterior is inherently permutation invariant, which provides a bias towards our disentanglement objective.

\textbf{Shuffle constraint }: Unfortunately, the numerous disentanglement biases included in this proposed architecture might not be sufficient to achieve a well-disentangled model. In order to achieve a clean disentanglement, it is therefore proposed to leverage the sequential nature of video data directly into the framework. Indeed, thanks to the aggregated posterior, the model directly provides the static code estimation for individual frames $q(\mathbf{f}_t|\mathbf{x}_t)$, without any additional computation. The idea is to condition the cNF, during training, on the shuffled $\mathbf{f}_{1:T}$ before aggregation, instead of conditioning the cNF on the aggregated static code $\mathbf{f}$. This is justified by the fact that, for a disentangled model, the $\mathbf{f}_{1:T}$ need to be invariant i.e., $p(\mathbf{x}_{1:T}|\mathbf{f})=p(\mathbf{x}_{1:T}|\mathbf{f}_{\pi(T)})$ with $\pi(T)$ denoting a random permutation.
Using this shuffled operation, the loss becomes: 
\begin{equation}
    \boxed{\mathcal{L} = \mathbb{E}_{p(\mathbf{x}_{1:T})}\mathbb{E}_{q(\mathbf{f}|\mathbf{x}_{1:T})}-\alpha \log p(\mathbf{x}_{1:T}|\mathbf{f}_{\pi(T)}) + \beta KL(q(\mathbf{f}|\mathbf{x}_{1:T})||p(\mathbf{f}))}
    \label{eq:final}
\end{equation}
which is the negative dataset ELBO where $p(\mathbf{x}_{1:T}|\mathbf{f})$ has been replaced by $p(\mathbf{x}_{1:T}|\mathbf{f}_{\pi(T)})$. Following the method of $\beta$-VAE \cite{higgins2016beta}, the different terms in $\mathcal{L}$ have additionaly been weighted using weight coefficients $\alpha$ and $\beta$. 

\textcolor{purple}{\textbf{Proposition 3.}} \textit{Consider any $\mathbf{h}$ and $q(\mathbf{f}|\mathbf{x}_{1:T})$ as described above which minimize $\mathcal{L}$ and further assume $\alpha\gg\beta$. Minimizing $\mathcal{L}$ is equivalent to minimizing both $L_{prop1}$ and $L_{propr2}$. Prop.1 and 2 are satisfied and the learned representation $(\mathbf{f},\mathbf{\lambda}_{1:T})$ is disentangled in the sense of Def.1.
}

The proof is given in App.A.4. Intuitively, shuffling the static codes $\mathbf{f}_{1:T}$ and using them to reconstruct the frames, enforces the static code to be time invariant, while simultaneously being informative due to the reconstruction objective (favored by imposing $\alpha\gg\beta$).  Meanwhile, because we minimize $-\log p(\lambda_{1:T})\approx H(\lambda_{1:T})$ through the cNF, the dynamic code is encouraged to be non informative. In turn, this leads to the minimization of both $L_{prop1}$ and $L_{prop2}$, resulting in a disentangled model. Further details are provided in App.C.2.

\textbf{Summary }: we showed that, through a simple shuffle of the static codes, the proposed model is provably disentangled without requiring any additional complex loss, unlike previous methods, with all disentanglement conditions directly satisfied by our novel evidence lower bound. This makes our method modality free and highly efficient by avoiding any unnecessary constraint. In the case where the factors are independent, the model tries to capture both $\mathbf{s}$ and $\mathbf{d}_{1:T}$. However,  the proposed model also generalizes to factorize the static and dynamic factors without assuming independence. It does so by directly modeling the causality and only capturing $\mathbf{\epsilon}_{d,1:T}$.

\textbf{Using pretrained Autoencoders }: Similarly to \cite{lippe2022citris}, to facilitate disentanglement, instead of directly extracting the two separate codes, it is proposed to first learn an Autoencoder trained to encode/decode the frames without disentanglement. Gaussian noise is added on the frame feature vectors to prevent collapse. After convergence, the autoencoder parameters are frozen and we learn the static encoder and conditional Normalizing Flow which provide the disentangled representation from the entangled one. Because a Normalizing Flow is invertible, it ensures no information is lost and the frozen decoder can be used to recover the frames without fine-tuning. This permits to fully factorize the disentanglement problem from the task of mapping high-dimensional complex images into a low dimensional space.

\section{Experiments}\label{sec:experiments}

In order to validate the effectiveness of the proposed model, it is compared qualitatively and quantitatively to state-of-the-art sequence disentanglement learning methods including MoCoGAN \cite{tulyakov2018mocogan}, DSVAE \cite{li2018disentangled}, R-WAE \cite{han2021disentangled}, S3VAE \cite{zhu2020s3vae}, SKD \cite{berman2023multifactor}, CDSVAE \cite{bai2021contrastively} and SPYL \cite{naiman2023sample}.

\textbf{Datasets }: Similarly to the baseline works, the model is evaluated using the standard \textit{LPCSprites} \cite{reed2015deep} and \textit{MUG} \cite{aifanti2010mug} databases, which contain videos of animated cartoon characters and facial expressions respectively. As all methods manage to provide high quality results on the original \textit{LPCSprites} dataset \cite{naiman2023sample}, it is modified in order to include more complex characters with static factors other than colors. Additionally, the \textit{MPI3D} database \cite{NEURIPS2019_d97d404b} is also considered in the experiments. It consists of a real-world robot arm carrying various objects in different positions with the arm motion made dependent on the background light, and the speed of the arm dependent on the object size. We further evaluate on the human action recognition \textit{MHAD} dataset \cite{chen2015utd} comprising videos of humans performing diverse actions. Finally, based on the \textit{dSprites} dataset \cite{dsprites17}, sequences of moving 2D shapes are generated. Two variants are proposed, one with fixed rotation and all factors independent (\textit{s-dSprites}) and the other with rotating shapes and objects that bounce on the edges according to their size (\textit{c-dSprites}). 

\textbf{Disentanglement metrics }: 
Quantitative evaluation is provided following the common benchmark for sequential disentanglement \cite{li2018disentangled,bai2021contrastively,zhu2020s3vae}. After encoding a sequence with known factors, either the static or dynamic code is replaced by a sample from its prior before reconstruction. A classifier, pretrained with full supervision, is then used to measure how well the factors associated with the fixed code are preserved for the reconstructed sequence. Four metrics are extracted : the accuracy for the fixed factors (Acc), the intra-entropy ($H(y|x)$) that measures the classifier confidence for the predictions, the inter-entropy ($H(y)$) that estimates the diversity for the sampled factors and the inception score (IS) that measures the generator performance.

Details on the model architecture and hyperparameters, datasets, disentanglement metrics as well as additional results can be found in App.B and C. 


\newcommand{\specialcell}[2][c]{%
  \begin{tabular}[#1]{@{}c@{}}#2\end{tabular}}

\begin{table*}[tb]
\caption{Disentanglement metrics on \textit{s-dSprites}, \textit{MUG} and \textit{MHAD} datasets with independent factors. The dynamic factors are highligted in \textcolor{red}{red}.}
\label{tab:table1}
\centering

 \begin{minipage}{.5\linewidth}
      \centering
      
         \begin{tabular}{c | c c c c} 
 \hline
  \textbf{MUG}&  \specialcell{Acc$\uparrow$ (\%) \\ \textcolor{red}{action}}  & IS$\uparrow$ & $H(y|x)$$\downarrow$ & $H(y)$$\uparrow$ \\[0.5ex] 
 \hline
  MoCoGAN &  63.12 & 4.332 & 0.183 & 1.721 \\ 
  DSVAE &  54.29 & 3.608 & 0.374 & 1.657\\
  R-WAE& 71.25 & 5.149 & 0.131& 1.771  \\
  S3VAE&  70.51 & 5.136 & 0.135 & 1.760 \\ 
  SKD&  77.45 & \textbf{5.569} & \textbf{0.052} & 1.769 \\ 
  C-DSVAE&  81.16 & 5.341 & 0.092 & 1.775 \\ 
  SPYL&  \textbf{85.71} & 5.548 & 0.066 & \textbf{1.779} \\ 
 \hline
 Ours & 84.32 & 5.329 & 0.068 & 1.778 \\ [1ex] 
 \end{tabular}
 
    \end{minipage}%
    \begin{minipage}{.56\linewidth}
      \centering

 \begin{tabular}{c | c c c c c c}  
 \hline
  & Acc$\uparrow$ & (\%) & &  IS$\uparrow$ \\ 
  \textbf{s-dSprites} &    color & shape & size &  \\ [0.5ex] 
 \hline
 DSVAE &  79.38 & 42.10  & 43.61 & 3.035  \\ 
 CDSVAE &  96.18 & 98.22 & 98.71 & 3.995  \\ 
 SPYL &  97.76 & 98.72 & 98.12 & 3.685  \\ 
 \hline
 Ours &  \textbf{98.91} & \textbf{98.98} & \textbf{98.76} & \textbf{4.263} \\  [1ex] 
 \end{tabular}

\begin{tabular}{c}
\quad\vspace{-1mm}
 \end{tabular}

 \begin{tabular}{c | c c c c} 
 \hline
  \textbf{MHAD}&  \specialcell{Acc$\uparrow$ (\%) \\ \textcolor{red}{action}}  & IS$\uparrow$ & $H(y|x)$$\downarrow$ & $H(y)$$\uparrow$ \\[0.5ex] 
 \hline
  C-DSVAE&  65.59 & 1.749 & 0.420 & 1.282 \\ 
  SPYL&  \textbf{68.63} & 1.780 & \textbf{0.362} & 1.299 \\ 
 \hline
 Ours & 67.65 & \textbf{1.896} & 0.371 & \textbf{1.325}\\ [1ex] 
 \end{tabular}
 
    \end{minipage} 

\end{table*}

\begin{table*}[tb]
\caption{Disentanglement metrics on \textit{c-dSprites}, \textit{MPI3D} and \textit{LPCSprites} with dependent factors. In \textcolor{red}{red} and \textcolor{blue}{blue} are highligted respectively the dynamic factors and the static factors that condition the motion.}
    \label{tab:table2}
    \centering

    \begin{tabular}{c | c c c c c c} 
 \hline
  & Acc$\uparrow$ & (\%) & &  IS$\uparrow$& $H(y|x)$$\downarrow$ & $H(y)$$\uparrow$\\ 
  \textbf{c-dSprites} &    color & \textcolor{blue}{shape} & \textcolor{blue}{size} &  &  &  \\ [0.5ex] 
 \hline
 DSVAE &  72.09 & 32.66 & 79.38 & 4.128 & 0.046 & 1.41 \\ 
 CDSVAE &  96.33 & 74.32 & 93.17 & 4.141 & 0.028 & 1.39 \\ 
 SPYL & 68.41 & 33.28 & 75.10 & \textbf{4.392} & 0.027 & 1.43 \\ 
 \hline
 Ours &  \textbf{98.92} & \textbf{98.98} & \textbf{98.59} & 4.371 & \textbf{0.008} & \textbf{1.44} \\  [1ex] 
 \end{tabular}

 \begin{tabular}{c}
\quad\vspace{-1mm}
 \end{tabular}

 \begin{tabular}{c | c c c c c c c c } 
 \hline
  & Acc$\uparrow$ & (\%) & & & IS$\uparrow$& $H(y|x)$$\downarrow$ & $H(y)$$\uparrow$\\ 
  \textbf{MPI3D} & shape & \textcolor{blue}{size} & \textcolor{blue}{camera} & \textcolor{blue}{light} &  &  &  \\ [0.5ex] 
 \hline
 DSVAE &  28.29 & 65.57 & 98.81 & 87.71 & 3.505 & 0.056 & 1.19 \\ 
 CDSVAE &  43.81 & 79.17 & \textbf{99.99}& \textbf{99.94}& 3.487 & 0.138 & 1.26 \\ 
 SPYL &  16.78 & 50.29 & 33.32& 86.54& 3.819 & 0.043 & 1.28 \\ 
 \hline
 Ours &  \textbf{90.15} & \textbf{97.56} & 99.97 & 99.93& \textbf{3.849} & \textbf{0.029} & \textbf{1.29} \\  [1ex] 
 \end{tabular}

 \begin{tabular}{c}
\quad\vspace{-1mm}
 \end{tabular}

\begin{tabular}{c | c c c c c c c c } 
 \hline
  & Acc$\uparrow$ & (\%) & & & IS$\uparrow$& $H(y|x)$$\downarrow$ & $H(y)$$\uparrow$\\ 
  \textbf{LPCSprites} & body  & hair & \textcolor{blue}{orient.} & \textcolor{red}{action} &  &  &  \\ [0.5ex] 
 \hline
 CDSVAE &  54.94 & 85.75 & 33.32& 67.71& 2.737 & 1.6e-3 & 1.00 \\ 
 SPYL &  70.89& 99.89 & 33.32& \textbf{99.86}& 2.735 & 2.0e-3 & \textbf{1.01} \\ 
 \hline
 Ours &  \textbf{99.97} & \textbf{99.91} & \textbf{99.94} & 99.83& \textbf{2.768} & \textbf{3.3e-4} & \textbf{1.01} \\  [1ex] 
 \end{tabular}

\end{table*}

\textbf{Results} :\label{sec:results}
The \textit{MUG}, \textit{MHAD} and \textit{s-dSprites} datasets have independent static-dynamic factors. Their corresponding results are shown in Tab.\ref{tab:table1}. As expected, when the factors are independent all recent methods manage to provide a disentangled representation. In particular, while using a simple shuffle operation, our proposed model performs on par with the complex state-of-the-art methods on all datasets. Despite the dynamic codes being extracted conditionally to $\mathbf{f}$, the learned representation correctly factorizes the independent factors. Only the SPYL method \cite{naiman2023sample} gives slightly better results on \textit{MUG} and \textit{MHAD}. However, we show in App.C.1 that this can be mainly explained by the tendency of SPYL to encode too much information in its dynamic codes. In Fig.\ref{fig:result2}, qualitative results for the MUG dataset are displayed with sequences generated by swapping the dynamic and static codes of test samples. The resulting videos accurately retain the person identities while exchanging the face motions with high quality non blurry outputs. 

For the \textit{LPCSprites}, \textit{c-dSprites} and \textit{MPI3D} datasets the static-dynamic factors are not independent. As an example, the \textit{c-dSprites} dataset displays 2D objects that move and rotate. Because the 2D shapes are either squares, ellipses or hearts, even if the rotation is generated independently, it is still causally influenced by the shape due to the different order of rotational symmetry between the shapes. 
Overall, this example highlights the fact that, even in simple cases, the static content of a scene may influence the dynamic of the sequence by conditioning the possible variations between frames. In Tab.\ref{tab:table2} the results for the three aforementioned datasets are given. Compared to the case with independent factors, the performance of the baseline methods significantly drop and those models fail to provide a disentangled representation. As explained in Sec.\ref{sec:background}, these methods extract the static and dynamic variables separately and enforce independence between the codes. This conflicts with the inherent causality of the ground truth data generative process (Sec.\ref{sec:formalism}) and results in ambiguity into where the shared information between the dynamic and static factors should lie. If the dynamic code is extracted independently, it will inevitably have to retain some static information, translating in entangled codes and potentially encouraging static factors into the dynamic variables. In contrast, our method manages to provide a disentangled representation and generalizes to settings with complex statistical relationships between the factors. The conditionality is properly captured and high accuracy scores are maintained. These results are illustrated in Fig.\ref{fig:result1} which shows sequences generated by fixing the static code and sampling the dynamic one. Compared to the competing CDSVAE model \cite{bai2021contrastively}, the designed method presents improved sample quality that maintains the object shapes and identity. The samples display diverse motions that abide by the causal mechanism (i.e., only vertical motion for the \textit{MPI3D} sample and fixed orientation for the \textit{LPCSprites} one, see App.B.1 for details on the datasets).

\textbf{Limitations} :
The proposed formalism (Sec.\ref{sec:formalism}) defines the factors. But, by doing so, it also draws the intrinsic limitations of the static/dynamic disentanglement problem as it provides what could be disentangled at best through DSVAE based methods. First, it is the dataset itself that defines what is static/dynamic, which might not match the human intuition or expectations, e.g., for the \textit{LPCSprites} dataset in Fig.\ref{fig:result1}, from a human perspective the orientation of a character would probably be labelled as a motion of that specific character identity. However, from the data perspective the orientation is static since it is time invariant. Second, the static code can only capture what is always invariant, i.e., factors that are static in every sequence of the dataset. A factor that can be either static or dynamic, depending on the sequence, is encompassed in $\mathbf{d}_{1:T}$. Hence, future work may investigate whether the proposed method can be extended to adaptively extract the static code for each sequence and further disentangle the factors at the sequence level.

\section{Conclusion}\label{sec:ccl}

In this work, we introduce a novel formal approach toward sequential data disentanglement. Diverging from previous works, it extends the VAE framework with a conditional Normalizing Flow that directly models the relationships between the static and dynamic factors, making it closer to the data generative process. This constitutes a crucial change compared to the baseline methods by lifting their most detrimental assumption. It translates into large improvements over the state-of-the-art models, which fail to disentangle dependent factors even in simple cases. To further enforce disentanglement, a simple shuffle constraint is proposed which leads to a novel ELBO formulation and our model to be provably disentangled. Hence, compared to the baselines, our method is simple, non dataset/domain specific and theoretically justified. It proved its ability to properly disentangle the static/dynamic factors on multiple datasets and to generalize to less restrictive cases with dependent variables. Interestingly, while our method has been only evaluated for videos, the presented approach directly extend to other domains such as audio or biology which could be explored in future works.

\section*{Acknowledgements}

Mathieu Cyrille Simon is a Research Fellow of the Fonds de la Recherche Scientifique - FNRS of Belgium. Computational resources have been provided by the supercomputing facilities of the Université catholique de Louvain (CISM/UCL) and the Consortium des Équipements de Calcul Intensif en Fédération Wallonie Bruxelles (CÉCI) funded by the Fonds de la Recherche Scientifique de Belgique (F.R.S.-FNRS) under convention 2.5020.11 and by the Walloon Region

%
%
\bibliographystyle{splncs04}
\bibliography{main}
\end{document}